\documentclass[runningheads]{llncs}
\usepackage[T1]{fontenc}
\usepackage{graphicx}
\begin{document}
\title{Hey AI Can You Grade My Essay?: Automatic Essay Grading}
%
%
\author{Maisha Maliha\inst{1} \and Vishal Pramanik\inst{2}}

\institute{
    School of Computer Science, University of Oklahoma, Norman, Oklahoma, United States.\\
    \email{maisha.maliha-1@ou.edu}
    \and
    Department of Computer Science and Engineering, Indian Institute of Technology Bombay, Mumbai, Maharashtra, India.\\
    \email{vishalpramanik35@gmail.com}
}

\maketitle
\begin{abstract}
Automatic essay grading (AEG) has attracted the
the attention of the NLP community because of its applications to several educational applications, such as scoring essays, short answers, etc.  AEG systems can save significant time and money when grading essays. In the existing works, the essays are graded where a single network is responsible for the whole process, which may be ineffective because a single network may not be able to learn all the features of a human-written essay. In this work, we have introduced a new model that outperforms the state-of-the-art models in the field of AEG. We have used the concept of collaborative and transfer learning, where one network will be responsible for checking the grammatical and structural features of the sentences of an essay while another network is responsible for scoring the overall idea present in the essay. These learnings are transferred to another network to score the essay. We also compared the performances of the different models mentioned in our work, and our proposed model has shown the highest accuracy of \textbf{85.50\%}.

\keywords{Automatic Essay Grading \and Collaborative learning \and Deep Learning \and Recursive Learning \and Recursive Neural Network.}
\end{abstract}
\section{Introduction}
Automatic essay grading (AEG) is a subfield of natural language processing (NLP) that has been around for more than a half-century.~\cite{1} suggested the first AEG system. For many years, little progress was made, mainly owing to a lack of resources (such as parsers) and processing power. But things changed with the development of deep neural networks and powerful computing machines.~\cite{2}, 
and ~\cite{3} are the best examples of applications of deep neural networks.

Automatic essay grading refers to using a machine to grade a text prepared in response to a topic known as the essay prompt. Holistic automatic essay grading refers to using computers to award an essay grade that reflects its overall quality. Aside from overall AEG, algorithms may be trained to rate essays depending on a single trait/attribute. This is referred to as trait-specific AEG. Content, organisation, and style are examples of essay characteristics. The necessity of human-graded essays submitted in response to the prompt in which the AEG system is evaluated is one of the problems in AEG. Domain adaption techniques such as cross-domain AEG have been developed to address this barrier. In our paper, we have contributed the following points:
\begin{itemize}
    \item We introduce a collaborative deep learning model for automatic essay grading. A model that not only looks at the nature and idea in the essay but also the grammatical and structural features of the sentences in the essay, before grading it.
    \item A comparative analysis of the results from different machine and deep learning models to our proposed deep learning network. 
\end{itemize}

\subsection{Difficulties in Automatic Essay Grading}
The most serious issue in the realm of AEG is \textbf{holistic AEG}. This procedure entails providing an overall score/grade to the essay depending on how well-written it is. The majority of AEG research is focused on holistic AEG. 

Much earlier research work used machine learning with classifiers to holistically grade the essays. Some commercial systems such as e-rater \textregistered ~\cite{4}, Intelligent Essay Assessor ~\cite{5}, etc. use a variety of features from surface-level features like length-based features (word count, average word length, average sentence length, etc.), to more complex features (like usage score, which detects usage errors in the essay) and kernels.

~\cite{6} describes a set of task-independent features for AEG, where they studied the problem of cross-domain AEG. One of their findings was that the best results for cross-domain AEG are when the source and target prompts are similar. 

\section{Related Works}
In this section, we look at different systems which are used for AEG.

\subsection{Classical Machine-Learning based Systems}
The earliest AEG systems use machine-learning-based approaches. The general approach for these systems involves feature engineering and ordinal classification/regression. Project Essay Grade (PEG) ~\cite{1}, the earliest AEG system, used a series of intrinsic properties (called \textit{trins}, which are analogous to features) to come up with an approximate score for the essay.

\subsection{Deep-Learning based Systems}
Since the early 2010s, neural networks have been used to solve multiple tasks in NLP, such as CNN ~\cite{7} and other hierarchical neural network models. ~\cite{8} develop a system which learns score-specific word embeddings (SSWE). They use these SSWEs, instead of other word embeddings like GloVe, to learn word embeddings before getting an essay representation via LSTMs to get the overall essay score. \cite{9} use LSTMs and other types of recurrent neural networks to score essays. They use a similar architecture as \cite{8}, except that they use pre-trained word embeddings.

\section{Background and Technologies}

\subsection{Recurrent Neural Networks}
A recurrent neural network (RNN)~\cite{15} is a deep neural network that uses the previous step's output as feedback input in the next step. They are generally used in sequences, in next-word prediction in a sentence. The RNNs store the knowledge in a hidden state and use the previous knowledge to predict the next sequence. At time step $t$, the hidden state $a^{<t>}$ and the output $y^{<t>}$ can be expressed as:
\begin{equation}
    a^{<t>}=g_{1}(W_{aa}a^{<t-1>} + W_{ax}x^{<t>} + b_{a})
\end{equation}
and
\begin{equation}
    y^{<t>}=g_{2}(W_{ya}a^{<t>} + b_{y})
\end{equation}
where $W_{aa}, W_{ax}, W_{ya}, b_{a}$ and $b_{y}$ are the coefficients that are shared over time steps and $g_{1}$ and $g_{2}$ are the activation functions. The working mechanism has been shown in the figure ~\ref{archi}.
\begin{figure*}[!h]
    \centering
   \fbox{\includegraphics[width=.80\columnwidth,height=0.35\columnwidth]{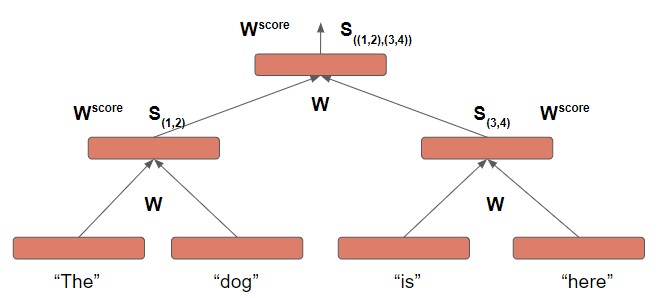}}
    \fbox{\includegraphics[width=.80\columnwidth, height=0.35\columnwidth]{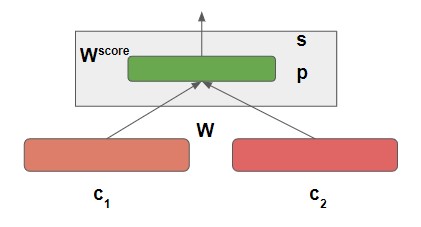}}
    \caption{The above images show how the RvNN works for a sentence. The word embeddings are fed to a neural network to get the phrase embedding, which is then passes into network to get the sentence embedding output vector. This vectors stores the grammatical and structural properties of the sentence}
    \label{archi}
\end{figure*}
\subsection{Convolutional Neural Network}
The Convolutional Neural Network~\cite{10} is a class of neural networks used for data processing with a grid-like structure, such as an image. A CNN has three main parts: \textbf{convolutional layer}, \textbf{pooling layer} and  \textbf{dense layer}. The convolution layer uses the concept of the kernel function to extract the information from the image when given as input in image matrix format.  The convolution can be calculated using the below equation.
\begin{equation}
    G[m,n]=(f*h)[m,n]=\sum_{j}\sum_{k}h[j,k]f[m-j,n-k]
\end{equation}
where $f$ and $h$ are the input images and the kernel function and $m$ and $n$ are the dimensions of the resultant image matrix and $j$ and $k$ are the dimensions of the kernel function. The pooling layer decreases the dimension of the input matrix and extracts the feature matrix's deeper meaning. 

\subsection{Bidirectional Encoder Representations from Transformers (BERT)}
BERT~\cite{11} is the Encoder~\cite{12} part of a Transformer ~\cite{13}. It is responsible for the extraction of features from the input sentence and pass the output vector to the Decoder part of the Transformer. BERT encode the input sentences with positional encodings so that the model can run in parallel. It consists of 12 encoder blocks, which contains a multi-headed self-attention mechanism and a dense layer. The attention mechanism helps to the different relations between the words in the sentences. This helps the model to have greater understanding of the sentences.

\section{Dataset}
We used the most commonly used AEG dataset, the Automated Students Assessment Prize (ASAP) Automatic Essay Grading (AEG) dataset, for our experimentation. The dataset comprises nearly 13,000 essays in response to 8 different essay prompts. The dataset is freely available and hosted on Kaggle, and the link is given in (https://www.kaggle.com/c/asap-aes/data). Overall, ASAP consists of eight essays, each corresponding to one question and was initially authored by students in grades 7 through 10. The statistics of the data has been given in the table \ref{t1}. The essays are scored based on four points: ideas, style, organization, and conventions.

\begin{table}
\begin{center}
\caption{Data Analysis of the Data showing the different prompts along with their number of essays, average length and score range}
\begin{tabular}{||c c c c||} 
 \hline
 Prompt & No. of Essays & Avg Length & Score Range \\ [0.5ex] 
 \hline\hline
 1 & 1783 & 350 & 2-12 \\ 
 \hline
 2 & 1800 & 350 & 1-6 \\
 \hline
 3 & 1726 & 150 & 0-3 \\
 \hline
 4 & 1772 & 150 & 0-3 \\
 \hline
 5 & 1805 & 150 & 0-4 \\  
 \hline
 6 & 1800 & 150 & 0-4 \\ 
 \hline
 7 & 1569 & 250 & 0-30 \\
 \hline
 8 & 723 & 650 & 0-60 \\  
 \hline
\end{tabular}
\label{t1}

\end{center}
\end{table}

The four types of essays present are persuasive, narrative, expository and source-dependent responses. We took the scores given by raters belonging to domain one and added the scores as the final score. We divided the dataset into 4:1 ratios for training and testing, respectively.

\section{Techniques and Implementation}
We have tested the dataset on the following models for comparative analysis.
\subsection{SVM}
We took the automatic grading problem as a classification task. As shown in Table 1, the maximum mark that can be obtained is 60, and the minimum mark is 0. The marks are divided into 61 classes from 0 to 60. We took the essay, used TF-IDF ( Term Frequency and Inverse Document Frequency )~\cite{14} vectorisation, and used 'l2' normalisation. The tf-idf does not blindly consider the number of times a word appears in the document by scaling down the words that appear frequently in the essays and have less meaning, such as the articles 'a','an' and 'the'. Suppose w is a word in the essay e then the tfidf is calculated by:
\begin{equation}
    tfidf(w,e) = tf(w,e) * idf(w) 
\end{equation}
where,
\begin{equation}
    idf(w) = log\frac{(1 + n)}{(1+df(w))} + 1
\end{equation}
as we smoothened the tfidf, $n$ is the number of essays and $df(w)$ is the document frequency of the word in the essay. We used the multi-class classification SVM with gaussian kernel. The model was trained on the essay set and finally evaluated. \cite{13} uses these kinds of machine learning models for their work.

\subsection{BERT}
We use the standard bert-base-cased model for automatic essay grading. The word "base" means it has 6 encoder blocks, and the word "cased" signifies that the model considers the case of the words while generating the result. We purposely took this model when grading an essay, the case of the words delivers important information to the model, like the beginning of a sentence or a proper noun in the sentence etc.   Because BERT cannot accept inputs of size greater than 512 tokens, we have eliminated the essays which have more words than 500 words. Each encoder block contains a self-attention layer followed by a feedforward layer. The attention layer contains 8 multi-headed attention mechanisms that store different information regarding a sentence's words. This information is used during the grading of the essay. The feedforward network maintains the dimension of the output vector, which is sent to the next encoder block. The positional encoding of the words with Bert embeddings makes the model run very fast, as it can process the essay parallelly.

\subsection{Collaborative Deep Learning Network (CDLN)}
Collaborative learning is a method of learning where the pupils work in groups and work on separate tasks contributing to a common overall outcome, or work together on a shared task. Collaborative learning models have the power of transfer learning, where instead of a single deep neural network to learn everything, the learning can be distributed among several networks, and their collective knowledge can be shared. Our model consists of a Recursive Neural Network, a Convolutional Neural Network, an LSTM and a dense neural network, as shown in the figure~\ref{archi}. The different parts of the architecture are described below in detail.

\begin{figure*}[!h]
    \centering
    \fbox{\includegraphics[width=1.0\columnwidth, height=0.55\columnwidth]{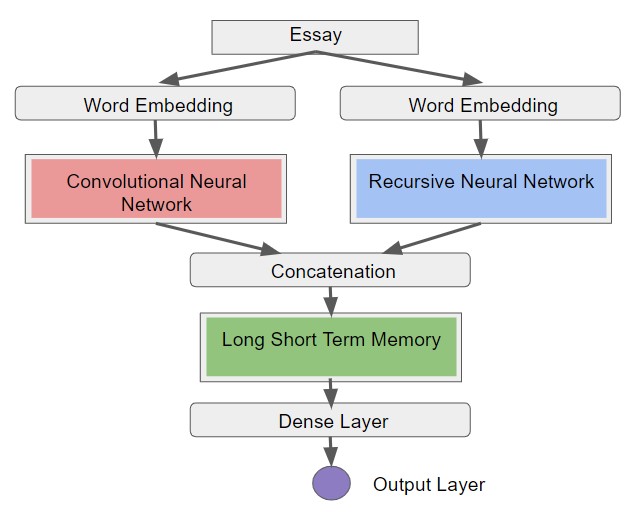}}
    \caption{The architecture of our Collaborative Deep Learning Network}
    \label{archi}
\end{figure*}
\subsubsection{Convolutional Neural Network}
The convolutional neural network understands the idea conveyed in the sentences using the convolution and the average pooling layers. The words of the essays are converted into Word2Vec embeddings of 100 dimensions each. All these vectors are concatenated together to form a single-dimension vector. This vector is sent forward for convolution and pooling. 1x105x8  kernel is used for convolution and  1x90x8 for average pooling. The convolution and pooling layer is repeated for 5 times. After this, the layer is flattened and sent forward for concatenation with the Recursive Neural Network (RvNN) output vector. The convolution layer helps to analyse the essay, while the average pooling layers bring out the idea conveyed in the sentences. 
\subsubsection{Recursive Neural Network}
The Recursive Neural Network (RvNN) ~\cite{15} understands the structure of the sentences of the essays. This helps the CDLN model check the essays' grammatical and sentence construction errors. The essays are first embedded with a word embedding, using the Word2Vec. The words are first divided into bigrams; then, the representation vectors are fed into a neural network as shown in figure~\ref{archi}.

We took the embedding dimension as 100. So the neural network has 200 neurons in the first layer. After that 4 layers of 150 neurons each. The output layer is 100 neurons to match up with word embedding. The formula is shown below.
\begin{equation}
    p = f(W[c_{1};c_{2}] + b) 
\end{equation}
\begin{equation}
    s = W^{score}p
\end{equation}
The $c_{1},c_{2}$ are the representation vectors, $W$ is a trainable vector, $b$ is the bias. $W^{score}$ is another trainable vector and $s$ is the score for that particular hierarchical structure of the sentence. Once the score is calculated, the same algorithm is followed for the rest of the sentences and then summed up and the final vector is forwarded for concatenation. An example of the algorithm with the sentence is given below.
\begin{flushleft}

\underline{\textbf{Algorithm}}\newline
Consider the sentence " The dog is here ", and each word is denoted by numerics '1', '2', '3' and '4', respectively and their embeddings by $c_{1}, c_{2}, c_{3}$ and $c_{4}$. We calculate the score using formulas 6 and 7.\\
\textbf{Step 1:} First, we calculate the scores for each word pair using the word embeddings of the words in the sentences given by s[1,2], s[2,3] and s[3,4] and store them in A[1,2], A[2,3] and A[3,4] respectively.\newline
\textbf{Step 2:} Then we use the scores in the previous steps, the scores s[1,3] and s[2,4]. A[1,3]=max(A[1,2]+$c_{3}$,A[2,3]+$c_{1}$) and A[2,4]=max(A[2,3]+$c_{4}$,A[3,4]+$c_{2}$)\newline
\textbf{Step 3:} The whole structure of the sentence is decided in the last step A[1,4]=max(A[1,3]+$c_{4}$, A[1,2]+A[3,4], A[2,4]+$c_{1}$)
\newline
\textbf{Step 4:} The score having the highest value represents the hierarchical structure of the sentence.
\newline

\end{flushleft}

\subsubsection{Long Short Term Memory}
The LSTM ~\cite{16} takes the input vector, which is the resultant of the concatenation of the vector output of the Recursive Neural Network and the CNN. The output vector of LSTM is 1x10000 dimensions. This is forwarded to the dense layer. LSTM gathers the learnings of the CNN and RvNN and stores the learning in its hidden cell. This is a good example of transfer learning, where the previous deep learning networks transfer their cumulative to the following deep neural network. It keeps knowledge not only about the sentence structure but also of the ideas conveyed in the essays. Also, this information is sent forwarded to the next layers for grading the essay.  

\subsubsection{Dense Layer and the Output Layer}
The dense layer takes the input from the output vector of the LSTM. It is 5 hidden layers with 120 neurons. The last output layer is a single cell which gives the essay grade.

\section{Experimentation}
We have experimented with 6 models, a) CDLN model, b)BERT, c)RNN, d)ANN and, e)SVM. The architecture of all the models has been described in the previous section. For all the models the learning rate is 0.0001 and the batch size for training is 32. For the CDLN model, BERT, RNN, ANN and SVM models, the number of epochs used is 15, 10, 8, 8, 8 and 6, respectively. Dropouts are used in the deep learning models to stop overfitting. In all the models during training, eight-fold cross-validation has been used. 

\section{Results and Analysis}
We use the mean square error (MSE), Pearson's Correlation Coefficient (PCC) and Quadratic Weighted Kappa (QWK).
\begin{itemize}
    \item Pearson's Correlation Coefficient~\cite{16}: It is generally used to measure the linear correlation or relationship between two variables. It is calculated over a range of -1 to 1, where 1 shows that the variables are highly correlated in the positive direction while -1 shows that they are not correlated; the two variables are in opposite directions. 
    \item Quadratic Weighted Kappa~\cite{17}: The most commonly used metric for measuring the performance of AEG systems is Cohen's Kappa with Quadratic Weights- i.e. Quadratic Weighted Kappa (QWK). We used this for our evaluation.
\end{itemize}
\subsection{Automatic Evaluation}
We compare the results from the different machine and deep learning models that has been mentioned in the above sections. The models' results we compare here are a)Our proposed CDLN model, b) LSTM, c) RNN, d) ANN, and e) SVM. We have also taken ~\cite{20} as the baseline model and compared the results of the TDNN(ALL), CNN-LSTM, CNN-LSTM-ATT and 2L-LSTM with our models. The results have been shown prompt-wise in table \ref{t3} and overall in table \ref{t4}. From the tables, it can be seen that our CDLN model outperforms the other models, including the baseline models. The sharing and transferring of knowledge from the CNN and the RvNN to the LSTM has boosted the results. The self-attention mechanism in BERT has led to better results than TDNN(Sem+Synt), the best-performing model in \cite{20}. The knowledge stored in the neurons of the deep networks has overcome the feature engineering process. If we go prompt-wise, the results mostly show that our models outperform the others.

\begin{table*}[htbp]
\centering
\caption{Performance Comparison Table for each Prompt among the CDLN model and the other baseline models}
\fbox{
\begin{tabular}{|c|c|c|c|}
\hline
&\multicolumn{3}{|c|}{\textbf{Prompt 1}} \\
\cline{2-4} 
\textbf{Models} & \textbf{\textit{Accu.}}& \textbf{\textit{PCC}}& \textbf{\textit{QWK}}\\
\hline
CDLN model& \textbf{.8853}&.7565& .7176\\
\hline
BERT & .8194& .7341& .6599\\
\hline
LSTM & .7506& .6649& .4260\\
\hline
 RNN & .7549& .6566& .4469\\
\hline
ANN & .7364& .6526& .4905\\
\hline
 SVM & .7449& .6865& .4998\\
\hline
2L-LSTM & -& .7375& .5128\\
\hline
CNN-LSTM & -& .6722& .6594\\
\hline
CNN-LSTM-ATT & -& .6531& .5348\\
\hline
TDNN(Sem+Synt) & -& \textbf{.7993}& \textbf{.7366}\\
\hline
\end{tabular}

\begin{tabular}{|c|c|c|c|}
\hline
&\multicolumn{3}{|c|}{\textbf{Prompt 2}} \\
\cline{2-4} 
\textbf{Models} & \textbf{\textit{Accu.}}& \textbf{\textit{PCC}}& \textbf{\textit{QWK}}\\
\hline
CDLN model& \textbf{.8466}& .7945& .7206\\
\hline
BERT & .8394& .7761& .6519\\
\hline
LSTM & .7524& .6765& .4295\\
\hline
 RNN & .7729& .6686& .4709\\
\hline
ANN & .7384& .6426& .4735\\
\hline
 SVM & .7499& .6895& .4798\\
\hline
2L-LSTM & -& .7375& .5128\\
\hline
CNN-LSTM & -& .6722& .6594\\
\hline
CNN-LSTM-ATT & -& .6531& .5348\\
\hline
TDNN(Sem+Synt) & -& \textbf{.7993}& \textbf{.7366}\\
\hline
\end{tabular}}

\fbox{
\begin{tabular}{|c|c|c|c|}
\hline
&\multicolumn{3}{|c|}{\textbf{Prompt 3}} \\
\cline{2-4} 
\textbf{Models} & \textbf{\textit{Accu.}}& \textbf{\textit{PCC}}& \textbf{\textit{QWK}}\\
\hline
CDLN model& \textbf{.8756}& \textbf{.7505}& \textbf{.7206}\\
\hline
BERT & .8394& .7761& .6519\\
\hline
LSTM & .7554& .6665& .4265\\
\hline
 RNN & .7729& .6686& .4709\\
\hline
ANN & .7454& .6596& .4456\\
\hline
 SVM & .6982& .6305& .4219\\
\hline
2L-LSTM & -& .6410& .5018\\
\hline
CNN-LSTM & -& .6844& .5454\\
\hline
CNN-LSTM-ATT & -& .5927& .4219\\
\hline
TDNN(Sem+Synt) & -& .6759& .6281\\
\hline
\end{tabular}

\begin{tabular}{|c|c|c|c|}
\hline
&\multicolumn{3}{|c|}{\textbf{Prompt 4}} \\
\cline{2-4} 
\textbf{Models} & \textbf{\textit{Accu.}}& \textbf{\textit{PCC}}& \textbf{\textit{QWK}}\\
\hline
CDLN model& \textbf{.8354}& \textbf{.7659}& \textbf{.7165}\\
\hline
BERT & .7950& .7509& .7364\\
\hline
LSTM & .7746& .6937& .6645\\
\hline
 RNN & .7246& .7067& .7146\\
\hline
ANN & .6765& .6864& .6594\\
\hline
 SVM & .5694& .5594 &.6044\\
\hline
2L-LSTM & -& .6527& .5754\\
\hline
CNN-LSTM & -& .7564& .7065\\
\hline
CNN-LSTM-ATT & -& .7224& .4665\\
\hline
TDNN(Sem+Synt) & -& .7616& .7578\\
\hline
\end{tabular}}

\fbox{
\begin{tabular}{|c|c|c|c|}
\hline
&\multicolumn{3}{|c|}{\textbf{Prompt 5}} \\
\cline{2-4} 
\textbf{Models} & \textbf{\textit{Accu.}}& \textbf{\textit{PCC}}& \textbf{\textit{QWK}}\\
\hline
CDLN model& \textbf{.8756}& .7505& \.7206\\
\hline
BERT & .8394& .7761& .6519\\
\hline
LSTM & .7554& .6665& .4265\\
\hline
 RNN & .7729& .6686& .4709\\
\hline
ANN & .7384& .6426& .4735\\
\hline
 SVM & .7049& .6365& .4298\\
\hline
2L-LSTM & -& .7375& .5128\\
\hline
CNN-LSTM & -& .6722& .6594\\
\hline
CNN-LSTM-ATT & -& .6531& .5348\\
\hline
TDNN(Sem+Synt) & -& \textbf{.7993}& \textbf{.7366}\\
\hline
\end{tabular}

\begin{tabular}{|c|c|c|c|}
\hline
&\multicolumn{3}{|c|}{\textbf{Prompt 6}} \\
\cline{2-4} 
\textbf{Models} & \textbf{\textit{Accu.}}& \textbf{\textit{PCC}}& \textbf{\textit{QWK}}\\
\hline
CDLN model& \textbf{.8561}& \textbf{.7573}& \textbf{.7094}\\
\hline
BERT & .8264& .7368& .6749\\
\hline
LSTM & .7654& .6809& .4275\\
\hline
 RNN & .7754& .6605& .4760\\
\hline
ANN & .7160& .6853& .5229\\
\hline
 SVM & .6489& .6195& .5699\\
\hline
2L-LSTM & -& .6528& .4951\\
\hline
CNN-LSTM & -& .6460& .5810\\
\hline
CNN-LSTM-ATT & -& .6291& .5149\\
\hline
TDNN(Sem+Synt) & -& .6903& .6752\\
\hline
\end{tabular}}

\fbox{
\begin{tabular}{|c|c|c|c|}
\hline
&\multicolumn{3}{|c|}{\textbf{Prompt 7}} \\
\cline{2-4} 
\textbf{Models} & \textbf{\textit{Accu.}}& \textbf{\textit{PCC}}& \textbf{\textit{QWK}}\\
\hline
CDLN model& \textbf{.8702}& \textbf{.7596}& \textbf{.7246}\\
\hline
BERT & .8349& .7719& .6537\\
\hline
LSTM & .7589& .6645& .4237\\
\hline
 RNN & .7749& .6659& .4721\\
\hline
ANN & .7379& .6464& .4754\\
\hline
 SVM & .7016& .6543& .4267\\
\hline
2L-LSTM & -& .7637& .6690\\
\hline
CNN-LSTM & -& .6849& .6609\\
\hline
CNN-LSTM-ATT & -& .6314& .6002\\
\hline
TDNN(Sem+Synt) & -& .7201& .6587\\
\hline
\end{tabular}

\begin{tabular}{|c|c|c|c|}
\hline
&\multicolumn{3}{|c|}{\textbf{Prompt 8}} \\
\cline{2-4} 
\textbf{Models} & \textbf{\textit{Accu.}}& \textbf{\textit{PCC}}& \textbf{\textit{QWK}}\\
\hline
CDLN model& \textbf{.8449}& \textbf{.7550}& \textbf{.7078}\\
\hline
BERT & .8252& .7578& .6904\\
\hline
LSTM & .7745& .6690& .4656\\
\hline
 RNN & .7606& .6603& .4584\\
\hline
ANN & .7454& .6596& .4456\\
\hline
 SVM & .6982& .6305& .4219\\
\hline
2L-LSTM & -& .5137& .2486\\
\hline
CNN-LSTM & -& .4666& .3812\\
\hline
CNN-LSTM-ATT & -& .5358& .4468\\
\hline
TDNN(Sem+Synt) & -& .6328& .5741\\
\hline
\end{tabular}}

\label{t3}

\end{table*}


\begin{table}[ht]
\centering
\caption{Overall Average Performance Comparison Table for all Prompt among the CDLN model and the other baseline models}
\begin{tabular}{|c|c|c|c|}
\hline
&\multicolumn{3}{|c|}{\textbf{Average}} \\
\cline{2-4} 
\textbf{Models} & \textbf{\textit{Accu. (\%)}}& \textbf{\textit{PCC}}& \textbf{\textit{QWK}}\\
\hline
CDLN model& \textbf{.8550}& \textbf{0.7545}& \textbf{0.7036}\\
\hline
BERT & .8268& .7390& .6779\\
\hline
LSTM & .7645& .6288& .4287\\
\hline
 RNN & .7786& .6645& .4712\\
\hline
ANN & .7305& .6456& .4787\\
\hline
 SVM & .7085& .6332& .4236\\
\hline
2L-LSTM & -& .6548& .4687\\
\hline
CNN-LSTM & -& .6569& .5362\\
\hline
CNN-LSTM-ATT & -& .6535& .5057\\
\hline
TDNN(Sem+Synt) & -& .7244& .6856\\
\hline
\end{tabular}
\label{t4}

\end{table}

\begin{figure}[htbp]
\centerline{\includegraphics[scale=0.65]{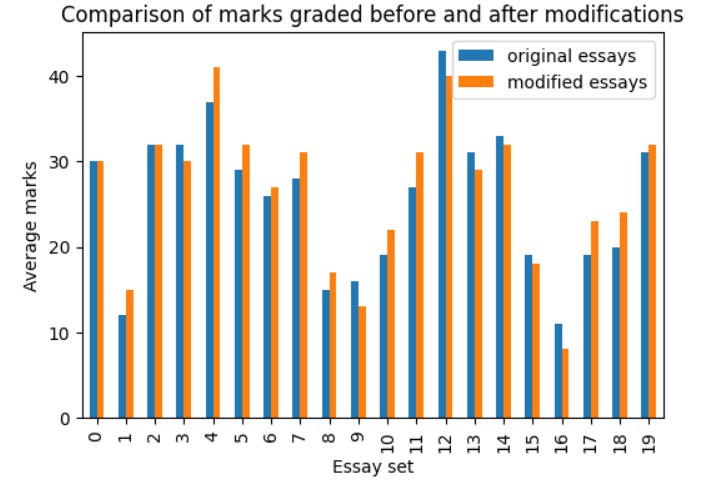}}
\caption{The image compares the average grade of the original essays and the essays after paraphrasing}
\label{comp}
\end{figure}

\subsection{Robustness of the model}
An automatic grading system must be robust in its action. It must grade the same essay with the same grade even when there is a slight change in the language and structure of the sentences. So we have conducted a robustness check of our CDLN model. For this test, we have selected 1000 random essays from the test set of all the essays. Then we rephrase the essays with Quillbot \footnote{https://quillbot.com/}, a well-known paraphrasing tool available freely for users. We graded the selected essays with our model. After this, we paraphrased the essays with Quillbot and graded them with our model. We recorded the grades and averaged the marks of every 50 essays and then plotted and compared them in the graph shown in figure ~\ref{comp}

It is evident from the above comparison-bar graph that the grades given by the model before and after paraphrasing is very close. It can be noticed from the graph that the average marks of the modified essays are more than the original essays; this can be due to the fact that quillbot has added proper sentence structure in the essays and removed the grammatical errors which may be present in the original essay. We further tested how much the grade differed from each other. For this purpose, we have modified the mean square error formula. If $g_{original}$ is the marks graded by the model for the original essays and $g_{modified}$ is the marks graded for the paraphrased essays then we find the difference by the formula:
\begin{equation}
    \Delta = \frac{\sum_{i=1}^{N}( g_{original} - g_{modified} )^{2}}{N}
\end{equation}
We used the above formula for the grades of all the $N=1000$ essays and got the value 0.34. This shows that grades are very close to each other, and the model is quite robust.

\section{Conclusion and Future Works}
Our study aims to demonstrate the power of collaborative learning in automatic essay grading problems. When multiple networks work together to understand and learn the different features of an essay and grade accordingly, their performance is boosted, as seen from the above table. This model has outperformed many of the pretrained and state-of-the-art models in AEG. But still, there are ample scopes of improvement in the results. With the introduction of many new deep neural networks and pretrained networks, these kinds of research can be further investigated and conducted. Our model gives a holistic score to the essay, not paragraph-wise. This can be further investigated.

%
%
%
%

\end{document}